\documentclass[runningheads]{llncs}
\usepackage{amsmath}
\usepackage{amsfonts}
\usepackage[T1]{fontenc}
\usepackage[misc]{ifsym}
\usepackage{graphicx}
\usepackage{subfigure}
\usepackage{hyperref}
\usepackage{color}

\usepackage{booktabs} 
\usepackage{float}

\begin{document}
\title{A Baseline Generative Probabilistic Model for Weakly Supervised Learning \thanks{Supported by organization JPMorgan Chase \& Co.}}
\toctitle{A Baseline Generative Probabilistic Model for Weakly Supervised Learning}
%
%
\author{Georgios Papadopoulos \Letter \orcidID{0009-0009-9732-3423} \and
Fran Silavong \and
Sean Moran}
\tocauthor{Georgios~Papadopoulos, Fran~Silavong, Sean~Moran}
\authorrunning{G. Papadopoulos et al.}
%
\institute{JPMorgan Chase \& Co., 25 Bank St, London E14 5JP, UK}
\maketitle              
\begin{abstract}
Finding relevant and high-quality datasets to train machine learning models is a major bottleneck for practitioners. Furthermore, to address ambitious real-world use-cases there is usually the requirement that the data come labelled with high-quality annotations that can facilitate the training of a supervised model. Manually labelling data with high-quality labels is generally a time-consuming and challenging task and often this turns out to be the bottleneck in a machine learning project. Weakly Supervised Learning (WSL) approaches have been developed to alleviate the annotation burden by offering an automatic way of assigning approximate labels (pseudo-labels) to unlabelled data based on heuristics, distant supervision and knowledge bases. We apply probabilistic generative latent variable models (PLVMs), trained on heuristic labelling representations of the original dataset, as an accurate, fast and cost-effective way to generate pseudo-labels. We show that the PLVMs achieve state-of-the-art performance across four datasets. For example, they achieve 22\% points higher F1 score than Snorkel in the \emph{class-imbalanced} Spouse dataset. PLVMs are plug-and-playable and are a drop-in replacement to existing WSL frameworks (e.g. Snorkel) or they can be used as baseline high-performance models for more complicated algorithms, giving practitioners a compelling accuracy boost.

\end{abstract}
\section{Introduction}
\label{intro}

In recent years, weakly supervised learning (WSL) has emerged as an area of increasing interest among machine learning practitioners and researchers. This interest has been driven by the need to automate the process of applying deep learning models to unlabelled real-world data, thus making manual annotations unnecessary and expensive. For example, medical doctors may wish to use machine learning (ML) models to improve the detection of intracranial haemorrhage (ICH) on head computed tomography (CT) scans \cite{Saab2019}, but current datasets are often large and unlabelled, making the application of ML the models difficult.

Various research teams, including Snorkel and Flying Squid \cite{Ratner2018a,Ratner2018b,Ratner2019,zhang2021}, have developed methods to address this labelling problem, with the overarching goal of reducing the cost of labelling for large datasets by hand. These WSL methods automate the otherwise tedious and costly manual labelling process by sourcing prior information from Subject Matter Experts (SMEs), which is used to create labelling functions $\lambda$ that are applied to the data. The output of this approach is typically a binary sparse matrix (labelling matrix) $\Lambda$.

Overall, the increasing interest in WSL reflects the potential of this approach to enable the more efficient and effective use of machine learning models on real-world data, even when labelled data is scarce or expensive to obtain. By leveraging SME guidance and prior knowledge, WSL methods offer a promising avenue for automating the labelling process, reducing costs, and enabling more widespread adoption of ML models in a range of applications.

We present a straightforward algorithm to create dichotomous classes on unlabelled datasets. Like \cite{Ratner2018a}, our method utilizes labelling functions $\lambda$ derived from Subject Matter Expert (SME) domain knowledge to programmatically annotate previously unlabelled data. The resulting annotations are represented as a labelling matrix $\Lambda$. Our approach relies on the assumption that the sparse input matrix $\Lambda$ contains sufficient information for robust model creation. Specifically, we propose to use a probabilistic generative latent variable model, Factor Analysis (FA), to map dependencies among the elements of the labelling matrix and generate a 1-dimensional latent factor $z$. We dichotomize the latent variable $z$ using the median and assign each group of observations to a binary class.

Our approach addresses the negative impact of class imbalance and label abstentions on existing WSL methods. We provide empirical evidence for the superior performance of the FA model compared to the state-of-the-art model, Snorkel, across three publicly available datasets and one internal curated dataset. We also compare the performance of FA with two more complex generative probabilistic latent variable models: Gaussian process latent variable models (GPLVM) with Sparse Variational Gaussian Processes (SVGP) and Variational Inference - Factor Analysis (VI-FA).

We show that FA as a WSL model outperforms other methods in Table \ref{tab:results_fa_snorkel}, where it achieved accuracy of 95\% for the source code classification task,  86\% in the YouTube Spam dataset, 86\% in the Spouse dataset, and 65\% in Goodreads dataset.

To summarise, the contribution of this paper is the following:

\begin{itemize}
\item \textbf{Impact of class imbalance:} We study the impact of class imbalance and label abstentions on existing WSL models \cite{Ratner2019}. This is not only an academic problem but also a common occurrence in real-world data and applied cases. We empirically illustrate this negative effect on three publicly available datasets, YouTube Spam dataset \cite{Alberto2015}, Spouse dataset \cite{Ratner2019}, and Goodreads dataset \cite{Wan2018,Wan2019}.
\item \textbf{Stronger performance:} As a solution, we propose to leverage FA for a new WSL method that outperforms current state-of-the-art models, including Snorkel \cite{Ratner2019}, as well as the benchmark probabilistic algorithms GPLVM-SVGP and VI-FA in terms of \emph{both} performance and resilience to class imbalance.
\item \textbf{Robustness and Causality:} We demonstrate the robustness of the proposed FA model under small datasets, class-imbalance and label abstentions. Also, it is proven that FA models offer causality between the labelling functions and the true labels \cite{Fabrigar1999}.
\item \textbf{Industrial Applicability:} We applied our method on internal data (JPMorgan) and evaluated our model in real-world cases (source code) by communicating with SMEs (firm engineers).We show our method scales well in industrial settings, is plug-and-play, and highly robust and accurate.
\end{itemize}

\section{Related Work}
\label{relatedwork}

WSL, as a research area, has become widely popular and has experienced a wealth of publications; with many culminating to end-to-end production systems \cite{Ratner2018b,Ratner2019}. Some real-world examples, from a diversified domain, which WSL methods have been applied, include healthcare \cite{Saab2019,Dunnmon2020,Saab2020,Fries2021,goswami2022}, human posturing and ergonomic studies \cite{Liu2021,zheng2021,bazavan2022}, multimedia and sound \cite{rao2021,manco2021,reddy2021}, dataset querying \cite{wolfson2021}, in business studies and behavioural analysis \cite{Mathew2021,jain2021,tseng2021}, and autonomous driving \cite{Weng2019}. 

In our paper, we draw motivation from recent research on data programming and matrix completion methods for WSL. Specifically, in \cite{Ratner2016} the authors use conditionally independent and user defined labelling functions with a probabilistic model optimised using the log-maximum likelihood and gradient descent methods. The true class label for a data point is modeled as a latent variable that generates the observed, noisy labels. After fitting the parameters of this generative model on unlabeled data, a distribution over the latent, true labels can then be inferred. \cite{Bach2017} expand the previous research by adding an \textbf{L1} regulariser to the \cite{Ratner2016}'s formula. The team created a first end-to-end programmatic pipeline by incorporating findings from the two previous papers, named Snorkel \cite{Ratner2018a,Ratner2018b,zhang2021}. They also replaced the sampling of the posterior from a graphical model with a matrix completion approach \cite{Ratner2018b}. 

The main shortcomings of the probabilistic approach that \cite{Ratner2016,Bach2017,Ratner2018b} are using are that is mathematically quite complex (for example the works of \cite{Ratner2016,Bach2017}). Also, challenging to implement as a plug-and-play solution on industrial scale projects. Finally, as we demonstrate in Sections \ref{modelformulation} and \ref{experiments}, fails to perform under class-imbalance and small datasets. 

One way to address the class imbalance performance problem and simplify the algorithms came from \cite{Varma2019}. The authors presented a structure learning method that relies on robust Principal Component Analysis (PCA) to estimate the dependencies among the different weakly supervision sources. They show that the dependency structure of generative models significantly affects the quality of the generated labels. This, thematically, is quite similar to our work. The main differentiating factor is that in \cite{Varma2019} they use PCA as a method to replace the lower rank and the sparse matrix from their previous work \cite{Ratner2018a}; whereas, we propose to use FA as the \emph{entire} WSL model. 

Our approach, compared to \cite{Varma2019}, allows users to plug-and-play any latent probabilistic models, without further modification. Another major difference is that our approach (FA) considers independent diagonal noise compared to spherical noise of the PCA, therefore as a model is better suited to map causality amongst the labelling functions ($\lambda$) and the ground truth \cite{Fabrigar1999}.

\section{Model Formulation}
\label{modelformulation}

\begin{figure}[!ht]
\vskip 0.2in
\begin{center}
\subfigure[YouTube Spam]{\includegraphics[scale=0.25]{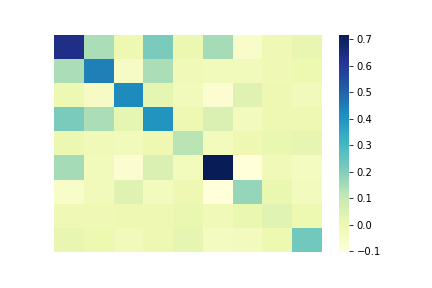}}
\subfigure[Spouse]{\includegraphics[scale=0.25]{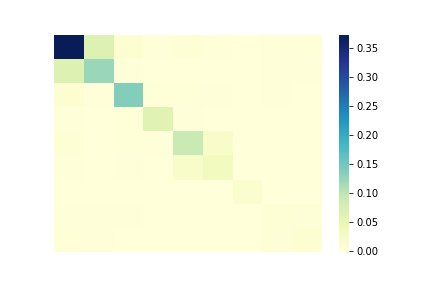}}
\subfigure[Goodreads]{\includegraphics[scale=0.25]{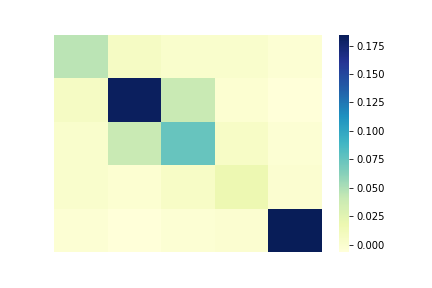}}
\caption{The covariance heatmaps of the three labelling matrices ($\Lambda$). Each heatmap is a dataset. We observe that the labelling functions $\lambda$ are independent. There is no strong relationship amongst them. But FA (and PLVM in general) are able to capture the causal relationship and approximate a true latent factor.}
\label{fig:covariance_matrix}
\end{center}
\vskip -0.2in
\end{figure}

In this paper we follow a two-step approach. Initially, we utilise heuristic labelling function techniques based on \cite{Ratner2018b} to create a sparse labelling matrix $\Lambda$. In the second step, we map the relationships (Fig. \ref{fig:covariance_matrix}) among the labelling functions using FA. Our approach has the benefit that can be expanded using any probabilistic generative latent variable model such as GPLVM.

\subsection{Labelling Functions}
\label{labellingfunctions}

Labelling functions $\lambda$, as described in \cite{Ratner2016}, are user-defined programmatic items that each incorporate the SME's knowledge in a binary form $\lambda \in \{0, 1\}$ or $\{-1\}$ if the function $\lambda$ is considered as abstain; where no relevant information is present. The goal of this process is to build a large set of approximate labels \cite{Bach2017}. Effectively, rather than hand-labelling training data, SMEs can write labelling functions instead. To this end, as a data programming approach, labelling functions offer model flexibility by programmatically expressing various weakly supervision sources, including patterns, heuristics, external knowledge bases, and more.

\subsection{Factor Analysis}
\label{factoranalysis}

Our objective is to enhance the generative methodology underlying Snorkel \cite{Ratner2019} with a more straightforward approaching. In a related work, \cite{Varma2019} applied a robust PCA to improve Snorkel results. However, our method differs from theirs in that they used PCA to initialize parameters for their probabilistic model, whereas we replace the scalable matrix completion algorithm with a probabilistic generative latent variable model (PLVM).

The use of generative latent models to extract underlying components from data has been extensively researched and documented \cite{Bartholomew1984,bishop2006,barber2012,Murphy2012}. A standard generative latent model is Factor Analysis (FA), which is closely related to Probabilistic PCA \cite{Tipping1999}. By leveraging the FA model, we aim to simplify the underlying probabilistic complexity of Snorkel and improve its performance on unlabelled datasets.

\subsection{Weakly Supervision with Factor Analysis}
\label{weakfactoranalysis}

Given an observed dataset $X \in \mathbb{R}^{n \times d}$, we utilise the labelling function $\lambda$ capabilities from Snorkel \cite{Ratner2019} to create a binary labelling matrix $\Lambda$. Labelling functions are user-defined programmatic items that scan the underlying data $X$ and result in the labelling matrix $\Lambda(X)$.  The labelling matrix is a $n{\times}m$ sparse matrix with $m$ the number of labelling functions $\lambda$, $n$ the number of data-points in the data $X$ and values $\Lambda \in \{0, 1, -1\}$. The Factor Analysis (FA) model captures the dominant dependencies amongst the data and subsequently finds a lower dimensional probabilistic description. FA can also be used for classification as they can model class conditional densities \cite{bishop2006}. In brief, the idea behind FA is that we have an observed dataset $\Lambda$ that is a linear representation of a latent factor $z$

\begin{equation} \label{eq:1}
\Lambda = Wz + c + \epsilon
\end{equation}

$W$ is the loading matrix with dimensions $m\times k$ with $k$ the dimensions of the latent factor $z$ with $k{\ll}d$, $c$ is a centred constant bias term and $\epsilon=\mathcal{N}(\epsilon|0,\Psi)$ is the Gaussian distributed noise of the model with $\Psi$ the $m{\times}m$ diagonal matrix. As a reminder $m$ is the number of columns/labelling functions in our observed data $\Lambda$ and $n$ the number of observed data-points. Probabilistically, this formula takes the form of the likelihood $p(\Lambda|z)$:

\begin{equation} \label{eq:2}
\begin{aligned}
p(\Lambda|z) & = \mathcal{N}(\Lambda|Wz + c, \Psi) \\
      & \propto exp(-\frac{1}{2}(\Lambda - W z - c)^{T}\Psi^{-1}(\Lambda - W z - c))
\end{aligned}
\end{equation}

the prior $p(z)$ of the Bayesian model is:

\begin{equation}
p(z) = \mathcal{N}(z|0, I) \propto exp(-\frac{1}{2}z^{T}z)
\end{equation}

This means that the centre of the factor $z$, due to its prior, will be constraint around 0. The next step to construct a full Bayesian model is to add the marginal $p(\Lambda)$:

\begin{equation}
p(\Lambda) = \int p(\Lambda|z)p(z)dz = \int \mathcal{N}(\Lambda|c, WW^{T} + \Psi) 
\end{equation}

The posterior $p(z|\Lambda)$:

\begin{equation} \label{eq:5}
\begin{aligned}
p(z|\Lambda) =&  \frac{p(\Lambda|z)p(z)}{p(\Lambda)} = \mathcal{N}(z|m, V) \\
& m = GW^{T}\Psi^{-1}(\Lambda - c) \\
& V = G + \mathbb{E}[z]\mathbb{E}[z]^{T}
\end{aligned}
\end{equation}

with $G = (I + W^{T}\Psi^{-1}W)^{-1}$.

The log-likelihood of this model is:

\begin{equation}\label{eq:6}
\begin{aligned}
\mathcal{L}(\Lambda|W, z, \Psi) &= -\frac{1}{2}trace((\Lambda - c)^{T}\Sigma^{-1}(\Lambda - c)) \\
      & - \frac{N}{2}log(2\pi) - \frac{1}{2}log|\Sigma|
\end{aligned}
\end{equation}

with $\Sigma = WW^{T} + \Psi$; where $\Psi$ the $m\times m$ noise diagonal matrix, $WW^{T}$ the $m\times m$ weights (loadings) matrix, and $\Sigma$ the $m\times m$ covariance matrix of the labelling data $\Lambda$.

Thus, the variance of the observed data ($\Lambda$) consists of a rank one component $WW^{T}$ originating from the joint dependence of the indicators on the latent factor $z$. Together with a full rank diagonal matrix $\Psi$, arising from the presence of noise, as it is an approximation of the latent variable. 

\subsection{Other Probabilistic Generative Latent Variable Models}
\label{benchmarkmodels}

In addition to Factor Analysis, we have also explored two alternative models in the family of PLVM. Specifically, we built a variational inference version of the Factor Analysis (VI-FA) using Tensorflow and the Adam optimiser, and we also put together a version of the GPLVM and SVGP models from GPflow.

\textbf{VI-FA model:} For this model, we followed a similar process as for the probabilistic PCA \cite{bishop2006} but using an independent variance for each data dimension $m$ (see Eq. \ref{eq:2}). To infer the posterior distribution of the latent variable model we utilise variational inference. We approximate the posterior $p(W, z, \Psi|\Lambda)$ (see Eq. \ref{eq:5}) using a variational distribution $q(W, z, \Psi)$ with parameters $\theta$. To find $\theta$ we minimise the KL divergence between the approximate distribution $q$ and the posterior, $KL(q(W, z, \Psi)|p(W, z, \Psi|\Lambda))$, which is to maximise the ELBO. 

\textbf{GPLVM model:} For the latter method, we trained a GPLVM model on the labelling matrix ($\Lambda$). By its nature, a GPLVM model can be interpreted as a generalisation of probabilistic PCA \cite{bishop2006}, where instead of optimising the linear mappings ($W$ in Eq. \ref{eq:1}) we optimise the latent variable $z$. In other words, it maps the connection between latent data $z$ and observable data $\Lambda$ using Gaussian-process priors. Overall, the log-likelihood from Eq. \ref{eq:6} becomes

\begin{equation}
\begin{aligned}
\mathcal{L}(\Lambda|W, z, K) &= -\frac{1}{2}tr((\Lambda - c) K^{-1} (\Lambda^{T} - c)) \\
      & - \frac{N}{2}log(2\pi) - \frac{1}{2}log|K| 
\end{aligned}
\end{equation}
with $K$ as the Gaussian process kernel.

During inference, the model accepts new (test) latent $z^{*}$-data and predicts the observable data $\Lambda^{*}$ by computing the two moments, mean and standard deviation. But, for our approach we need to be able to accept new observable data $\Lambda^{*}$ and predict the latent $z^{*}$-data. Similar to any other non-linear kernel-based model it is difficult for the GPLVM to be used as a dimensionality reduction tool that accepts test data. This is because it is challenging to invert the mapping between $z$ and observable $X$ (or $\Lambda$ in our case). Various approaches have been proposed that involve learned pre-images and auxiliary models \cite{Weston2004,Lawrence2006,Dai2016}. 

After training the GPLVM (Radial Basis Function kernel), we use an auxiliary Bernoulli regression model (SVGP) with Matern52 kernel to create the mapping between the latent target variable $z$ and the covariates of the regression model $\Lambda$. Then, for new data $\Lambda^{*}$ we use the SVGP model for predicting $z^{*}$.

\section{Datasets}
\label{datasets}
In this section, we describe the \emph{four} datasets used to evaluate the model performance between Snorkel and PLVMs. Three of them are publicly available and commonly used in the field of weakly supervised learning, and one is internally sourced. Table \ref{tab:datasets} provides the summary statistics.

\begin{table}[H] 
\caption{Dataset Statistics. $\lambda$ is the labelling function. Absent, shows the number of rows $n$ in the labelling matrix $\Lambda$ that have all the columns $m$ assigned as absent $\{-1\}$. \textbf{*}For the Spouse dataset we do not have the target values for the training data, only for the test sub-set. In the table we use the test data information. For the training data ($n=22,254$) the number of absent rows is $n=16,520$ or 74\%.}
\label{tab:datasets}
\vskip 0.15in
\centering
\resizebox{0.48\textwidth}{!}{%
\begin{tabular}{@{}llcccc@{}}
\toprule
& & \multicolumn{2}{c}{Number of} \\
 &  & Positive & Negative & Absent & $\lambda$ \\ \midrule
Source Code & Balanced & 127 & 123 & 0 & 3\\
Spam & Balanced & 831 & 755 & 230 & 9\\
Spouse & Unbalanced\textbf{*} & 218 & 2,483 & 1,951 & 9\\
Goodreads & Unbalanced & 514,778 & 281,293 & 691,795 & 5\\ \bottomrule
\end{tabular}%
}
\vskip -0.1in
\end{table}

\textbf{YouTube Spam Comments:} We use YouTube comments dataset, originally introduced in \cite{Alberto2015}. The comments were collected via the YouTube API from five of the ten most viewed videos on YouTube in the first half of 2015. The training data have $n=1,586$ YouTube video messages and the test data size is $n=250$. \cite{Ratner2019} created the labelling functions that include 5 keyword-based, 1 regular expression-based, 1 heuristic, 1 complex preprocessors, and 2 third-party model rules.

\textbf{Spouse Dataset:} This dataset is constructed by \cite{Ratner2019} to identify mentions of spouse relationships in a set of news articles from the Signal Media. The data is split between $n=22,254$ training samples and $n=2,701$ testing samples. There are 9 heuristic and NLP related labelling functions \footnote{\url{https://github.com/snorkel-team/snorkel-tutorials/blob/master/spouse/spouse_demo.ipynb}}. The ground truth labels for the training set are not available. Therefore, we are unable to check for class imbalance or the accuracy of the model on the training set.

\textbf{Goodreads Dataset:} We use the Goodreads dataset, from \cite{Wan2018} and \cite{Wan2019}. This data is a smaller sample from the original dataset and contains $n=794,294$ training records and $n=44,336$ test records, collected from $876,145$ Goodreads' users (with detailed meta-data). We followed the same experiment settings\footnote{\url{https://github.com/snorkel-team/snorkel-tutorials/blob/master/recsys/recsys\_tutorial.ipynb}} defined by Snorkel, where the task is to predict whether a user favours the book or not given the interaction and metadata as context.

\textbf{Source Code Dataset:} In addition to the natural language based tasks, we have also created a pipeline and evaluated our proposed method in an industrial setting at JPMorgan; on a binary classification task in the field of Machine Learning on Source Code (MLonCode). The objective was to predict the label of each function/method within a set of source code repositories. To the best of our knowledge, this is the first attempt of applying weakly supervised learning on source code. We internally curated $n=250$ functions and asked experienced senior software engineers to construct three labelling functions. The three labelling functions represent empirical methods that the engineers would have used if they were to manually assess the quality of the code of the function/method. This results in a class balanced source code dataset as indicated in Table \ref{tab:datasets}.

\section{Experiments}
\label{experiments}

Our aim is to validate the three main hypothesis of the paper: 1) the factor analysis model can be used for binary classification tasks; 2) the labelling matrix that contains the observable variables ($\Lambda$) of the model is the sufficient statistics of the model; 3) using PLVMs we achieve better results compared to existing methodologies. We ran our experiments using the following configurations: MacBook Pro 2019, Python 3.7.10, Snorkel 0.9.7, Sklearn 1.7.0, Tensorflow 2.6.0, Tensorflow-probability 0.13.0. The FA method that is used in Sklearn follows the SVD approach from \cite[p. 448]{barber2012}. For the alternative models that we used, Variational Inference Factor Analysis (VI-FA) and Gaussian process latent variable models - Sparse Variational Gaussian process (GPLVM-SVGP), we relied on Tensorflow and GPflow 2.2.1. All our models and data shuffling were set with random key $\{123\}$.

\begin{figure}[!ht]
\vskip 0.2in
\begin{center}
\subfigure[YouTube Spam]{\includegraphics[scale=0.25]{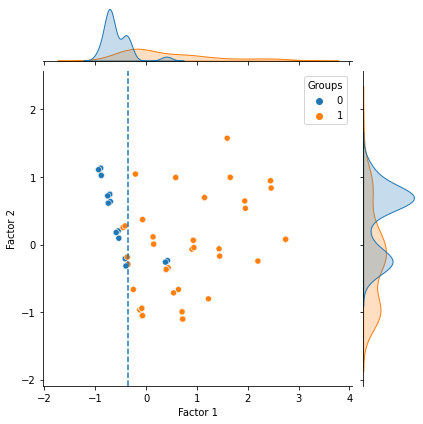}}
\subfigure[Spouse]{\includegraphics[scale=0.25]{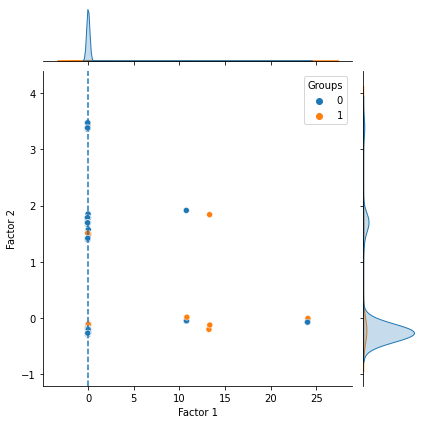}}
\subfigure[Goodreads]{\includegraphics[scale=0.25]{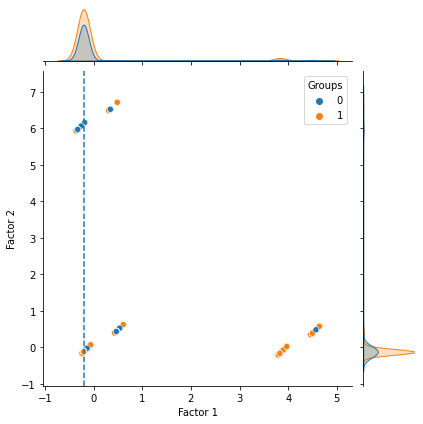}}
\caption{Jointplots between the two factors from the FA model. The top and right sides of the plots illustrate the distribution of each factor for the respective test set. The dotted line shows the median of the first factor, which is the one we used to dichotomise it and infer the classes during inference. The median is calculated on the training data factor. We observe that utilizing the first factor is sufficient to separate the two labels (blue and orange).}
\label{fig:joinplots}
\end{center}
\vskip -0.2in

\end{figure}

\textbf{Selection of classification threshold:} To create binary values from the latent factor $z$ we tested a series of thresholding methods. The best approach was to dichotomise the normally distributed test factor $z*$ with a median computed from the training factor $z$. Figure \ref{fig:joinplots} shows that the first factor's median (dotted line) of the YouTube Spam data divides the two separate groups accurately. For the Spouse and the Goodreads datasets, the median threshold again separates the two groups but not as effectively as in the previous case. Table \ref{tab:thresholds} shows the performance of using median, mean and Youden's J statistics as the threshold, where median achieves the best results. Thus, we propose to use median as the thresholding method.

\begin{table}[H] 
\caption{Threshold selection. The table shows the accuracy scores for each threshold choice. The CDF threshold was calculated using the Youden's J statistic after we transformed the test $z*$ with the normal CDF.}
\label{tab:thresholds}
\vskip 0.15in
\centering
\begin{tabular}{@{}lcccc@{}}
\toprule
    & & Median & Mean & CDF  \\ \midrule
Source Code & & 0.95 & 0.92 & 0.95 \\
Spam & & 0.86 & 0.74 & 0.85 \\
Spouse & & 0.86 & 0.92 & 0.90 \\
Goodreads & & 0.63 & 0.39 & 0.62 \\ \bottomrule
\end{tabular}
\vskip -0.1in
\end{table}

\textbf{YouTube Spam comments:} The YouTube Spam data \cite{Alberto2015} is a balanced dataset, with positive class numbers (1) $n=831$ and negative class numbers (0) $n=755$ (Table \ref{tab:datasets}). The Snorkel and FA models achieve close results, with accuracy $86\%$ for both methods, precision $83\%$ for Snorkel and $85\%$ for FA, recall $88\%$ and $84\%$, and F1 score $86\%$ and $85\%$. 

\begin{table*}[t] 
\caption{Accuracy, precision, recall, and F1 metrics for Source Code Classification, YouTube spam, spouse and Goodreads datasets, comparing the Snorkel approach against the FA WSL model. Each model is trained on the training dataset and evaluated on the test set. Bold numbers indicates the best performance.}
\label{tab:results_fa_snorkel}
\vskip 0.15in
\resizebox{\textwidth}{!}{%
\begin{tabular}{@{}lcccccccc@{}}
\toprule
 & \multicolumn{2}{c}{\textbf{Source Code}} & \multicolumn{2}{c}{\textbf{Spam   Dataset (NLP)}} & \multicolumn{2}{c}{\textbf{Spouse   Dataset (NLP)}} & \multicolumn{2}{c}{\textbf{Goodreads   Dataset (Recommender systems)}} \\ \midrule
 & \textbf{Snorkel} & \textbf{Factor Analysis} & \textbf{Snorkel} & \textbf{Factor Analysis} & \textbf{Snorkel} & \textbf{Factor Analysis} & \textbf{Snorkel} & \textbf{Factor Analysis} \\ \midrule
\multicolumn{1}{l}{\textbf{Accuracy}} & 0.92 & \textbf{0.95} & \textbf{0.86} & \textbf{0.86} & 0.54 & \textbf{0.86} & 0.53 & \textbf{0.63} \\
\multicolumn{1}{l}{\textbf{Precision}} & 0.90 & \textbf{0.97} & 0.83 & \textbf{0.85} & 0.12 & \textbf{0.32} & \textbf{0.66} & 0.65 \\
\multicolumn{1}{l}{\textbf{Recall}} & \textbf{0.95} & 0.93 & \textbf{0.88} & 0.84 & \textbf{0.72} & 0.64 & 0.56 & \textbf{0.95} \\
\multicolumn{1}{l}{\textbf{F1}} & 0.93 & \textbf{0.95} & \textbf{0.86} & 0.85 & 0.20 & \textbf{0.42} & 0.61 & \textbf{0.77} \\ \bottomrule
\end{tabular}%
}
\vskip -0.1in
\end{table*}

\begin{table*}[t]
\caption{Accuracy, precision, recall, and F1 metrics for Source Code Classification, YouTube spam, spouse and Goodreads datasets, comparing the FA WSL model against the VI-FA and GPLVM-SVGP models. The performance has been measured on the test sample of each dataset. Bold numbers indicates the best performance.}
\label{tab:results_fa_gplvm}
\vskip 0.15in
\resizebox{\textwidth}{!}{%
\begin{tabular}{@{}lcccccccccccc@{}}
\toprule
\textbf{} & \multicolumn{3}{c}{\textbf{Source Code Classification}} & \multicolumn{3}{c}{\textbf{Spam Dataset (NLP)}} & \multicolumn{3}{c}{\textbf{Spouse Dataset (NLP)}} & \multicolumn{3}{c}{\textbf{Goodreads Dataset   (Recommender systems)}} \\ \midrule
\textbf{} & \textbf{Factor Analysis} & \textbf{VI-FA} & \textbf{GPLVM} & \textbf{Factor Analysis} & \textbf{VI-FA} & \textbf{GPLVM} & \textbf{Factor Analysis} & \textbf{VI-FA} & \textbf{GPLVM} & \textbf{Factor Analysis} & \textbf{VI-FA} & \textbf{GPLVM} \\ \midrule
\textbf{Accuracy} & \textbf{0.95} & 0.92 & 0.92 & \textbf{0.86} & 0.70 & 0.82 & \textbf{0.86} & 0.78 & 0.09 & \textbf{0.63} & 0.60 & \textbf{0.63} \\
\textbf{Precision} & \textbf{0.97} & 0.89 & 0.89 & 0.85 & \textbf{0.89} & 0.79 & \textbf{0.32} & 0.22 & 0.08 & \textbf{0.65} & 0.64 & 0.64 \\
\textbf{Recall} & 0.93 & \textbf{0.95} & \textbf{0.95} & 0.84 & 0.42 & \textbf{0.85} & 0.64 & 0.70 & \textbf{0.99} & 0.95 & 0.89 & \textbf{0.96} \\
\textbf{F1} & \textbf{0.95} & 0.92 & 0.92 & \textbf{0.85} & 0.58 & 0.82 & \textbf{0.42} & 0.34 & 0.15 & \textbf{0.77} & 0.74 & \textbf{0.77} \\ \bottomrule
\end{tabular}%
}
\vskip -0.1in
\end{table*}

\textbf{Spouse Dataset:} In the Spouse dataset \cite{Ratner2019}, the FA achieves much higher performance compared to the Snorkel model. The dataset suffers from \emph{severe} class-imbalance and a large number of absent labelled classes, as shown in Table \ref{tab:datasets}. Specifically, from the $n=22,254$ training data, 74\% or $n=16,520$ observations have absent values ($\lambda=-1$) in all $m=9$ labelling functions. 

The FA model shows its strength on this type of dataset that have high number of absent items and experience class-imbalance. In terms of accuracy, Snorkel scores 54\%, whereas our model attains an impressive 86\%. On the other hand, in recall, Snorkel shows a score of 72\% and our model 
64\% (Table \ref{tab:results_fa_snorkel}). 

\textbf{Goodreads Dataset:} This is the largest dataset we used for model training and predictions \cite{Wan2018,Wan2019}. Similarly to Spouse data, Goodreads is a class-imbalanced dataset and it exhibits a considerable amount (87\% of the observations) of absent labelled items (Table \ref{tab:datasets}). In Table \ref{tab:results_fa_snorkel}, FA beats Snorkel predictions on almost every classification metric, namely 10\%+ accuracy, 1.69x recall and 16\% higher F1 score, and achieves marginally lower precision (1\%).

\textbf{Source Code Classification:} Table \ref{tab:results_fa_snorkel} shows the classification performance of Snorkel and FA when evaluated against the truth data. Specifically, the accuracy is $92\%$ for Snorkel and $95\%$ for the FA model; the precision is $90\%$ for Snorkel and $97\%$ for the FA; recall is $95\%$ and $93\%$ subsequently; and finally the F1 score is $93\%$ for Snorkel and $95\%$ for the FA.

\begin{figure}[!ht]
\vskip 0.2in
\begin{center}
\includegraphics[width=0.45\textwidth]{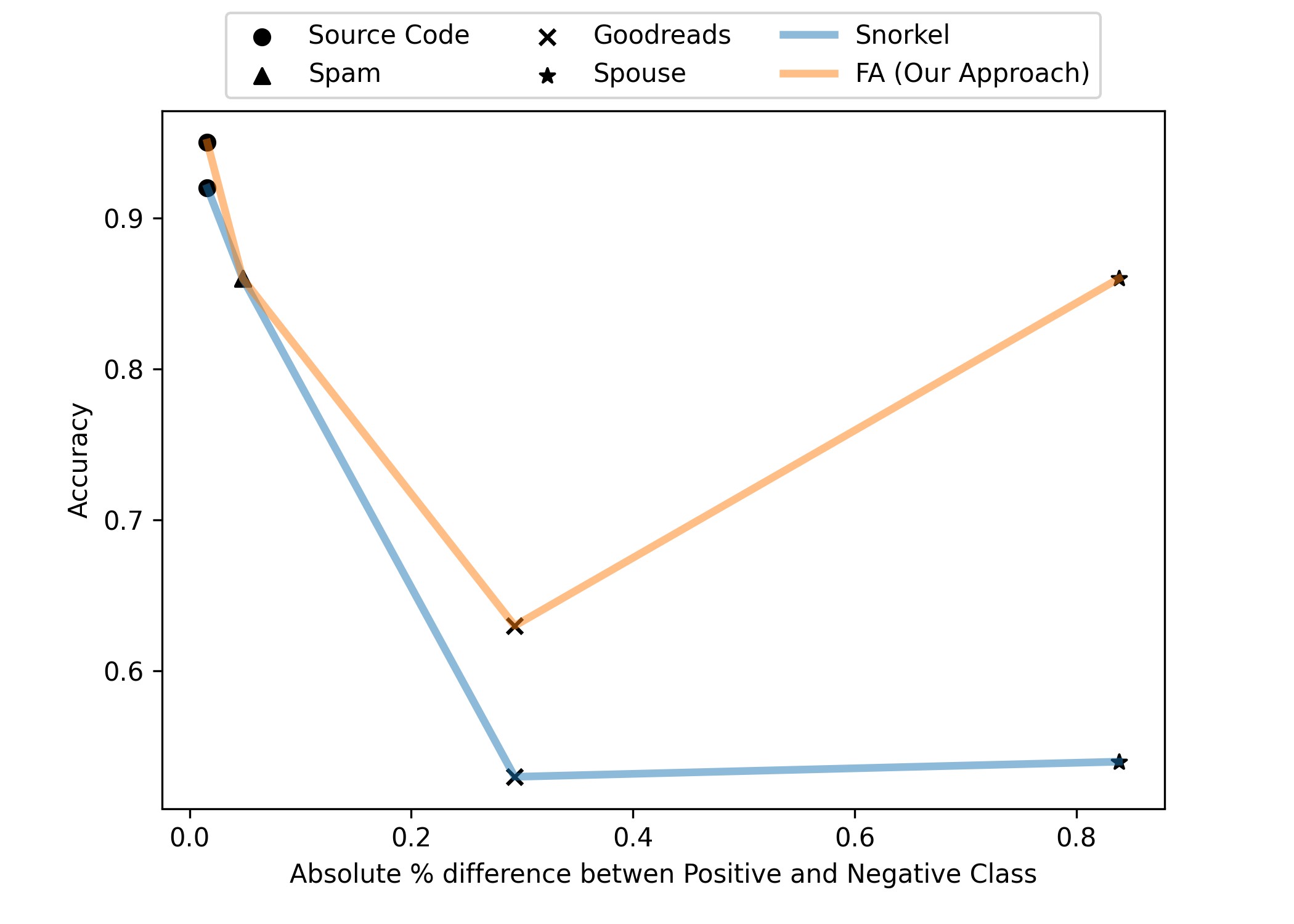}
\caption{Relation between the accuracy and absolute percentage difference of positive and negative classes representing the extent of class imbalance. We see evidence that the accuracy of Snorkel decreases with the extent of class imbalance, whereas Factor Analysis does not. FA also consistently outperforms Snorkel, when we compare the model accuracy for binary classification tasks on the four datasets: Youtube Spam (triangle), Spouse (star), Goodreads (cross), Internal Source Code (circle) and the extent of class imbalance.}
\label{fig:class_imbalance}
\end{center}
\vskip -0.2in
\end{figure}

\textbf{Class Imbalance and Abstentions:} To examine the relationship between model performance and class imbalance, we first quantify the class imbalance by computing the absolute percentage difference between positive and negative, $\frac{|{n_{pos} - n_{neg}}|}{n_{pos} + n_{neg}}$ of each dataset, where $n_{pos}$ and $n_{neg}$ refer to the number of positive and negative class respectively. We then compared this to the result stated in Table \ref{tab:results_fa_snorkel}. In Figure \ref{fig:class_imbalance}, we observe promising evidence to suggest that as the extent of class imbalance increases, the performance of Snorkel drops, whereas Factor Analysis model does not. 

The effect of class imbalance and abstentions can also be viewed in Figure \ref{fig:joinplots}. In the Spouse \emph{(a)} and in the Goodreads \emph{(b)} figures, we observe that abstentions make more challenging for the model to dichotomise efficiently the classes. Nevertheless, the FA method performs much more accurately compared to Snorkel (Table \ref{tab:results_fa_snorkel}) or the other two PLVM models (Table \ref{tab:results_fa_gplvm}). 

In general, abstentions and class-imbalance are two critical issues when we build a WSL pipeline. The probabilistic mechanism of FA, and how it maps the dependencies across the functions in the labelling matrix $\Lambda$, weaken the impact of these two problems significantly. 

\begin{figure}[!ht]
\vskip 0.2in
\begin{center}
\subfigure[YouTube Spam]{\includegraphics[scale=0.25]{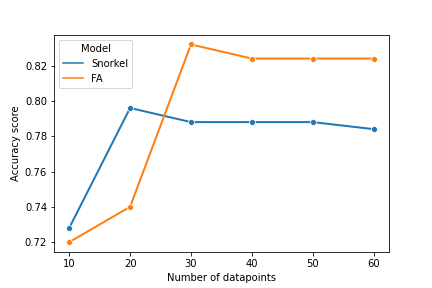}}
\subfigure[Spouse]{\includegraphics[scale=0.25]{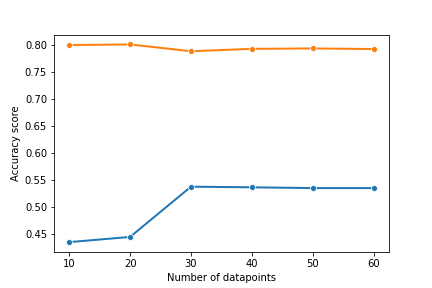}}
\subfigure[Goodreads]{\includegraphics[scale=0.25]{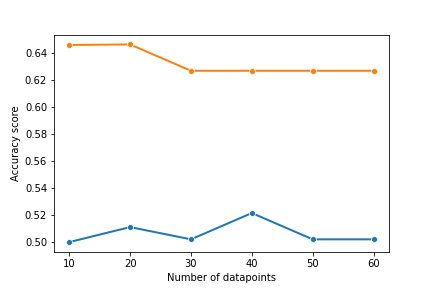}}
\caption{Comparison of classification performance of Snorkel and Factor Analysis (our proposed approach) in terms of accuracy. We randomly selected $n$ sample as the training set forming the labelling matrix $\Lambda$, where $n \in \{10, 20, 30, 40, 50, 60\}$ and evaluated against their respective test set. Factor Analysis achieves significant higher accuracy than Snorkel with merely 10 training samples in the Spouse and Goodreads datasets, and 30 training samples for the YouTube Spam dataset illustrating the robustness of our approach.}
\label{fig:lineplots}
\end{center}
\vskip -0.2in
\end{figure}

\textbf{Robustness:} We studied how the model performance changes when we vary the size of the training data. Figure \ref{fig:lineplots} shows that Factor Analysis achieves a $1.7x$ and 14\% higher accuracy in Spouse and Goodreads test set with only 10 training datapoints, and 4\% higher in YouTube Spam test set with 30 training datapoints.

\section{Benefits of the model}
\label{benefits}

The benefits of using the benchmark model to replace the pipeline of the Snorkel algorithm, include the speed, the robustness of the results on unbalanced datasets, the causality that FA offers \cite{Fabrigar1999}, and the explainability of the model compared to the Snorkel probabilistic approach. Explainability, as in terms of the FA model is a method that has been tested, evaluated, and used for years in the field of social sciences and the underlying mechanisms have been studied extensively. Finally, as we demonstrated in the evaluation of source code in JPMorgan, our approach can be easily integrated into existing machine learning workflows using standard widely-used libraries in Python (scikit) or R (psych). This allows users to leverage their existing knowledge and resources to quickly adopt and integrate the new framework into their applications.

\section{Limitations}
\label{limitations}

While the proposed method offers several benefits, there are some limitations that require further investigation in future research. The model's inability to perform on multi-label datasets, which are commonly encountered in many real-world applications. Multi-label datasets involve instances that can be assigned multiple labels or categories simultaneously, making them more complex than single-label datasets. Unfortunately, the model developed in this research was not able to effectively handle this type of data. We attempted to address this issue by increasing the number of principal components in the model output, but this did not yield significant improvements in performance.

\section{Discussion and Future Direction}
\label{discussion}

We introduced generative probabilistic latent variable models (Factor Analysis) as a novel approach to solve weakly supervised learning tasks. Our method, by using high-level domain knowledge from Subject Matter Experts, accomplishes high quality results and can be an excellent choice for approximating true labels on unlabelled data. We provided evidence that Factor Analysis is resilient to class imbalanced datasets as indicated by the significant improvement to the classification performance. Finally, we tested the effect of sparse data resources by varying the number of data-points used to train the generative model and we showed that with a minimum number of points our approach can attain high performance. For future work, we hope to expand the generative probabilistic latent variable models into a multi-class domain and explore our approach to other weakly supervised learning tasks.

\subsubsection{Acknowledgements} We want to thank the reviewers for their time, effort, and the very constructive feedback and advice. Our aim was to try and incorporate as many of their suggestions as possible considering the time. If some of their suggestions are not present (e.g. more datasets) is purely because of the limited timeframe. The readers can find the relevant code, as soon as it becomes available, at the JPMorgan Github \url{https://github.com/jpmorganchase} under the repository name \textit{weakly-supervision} and the branch \textit{ecml-experiments}. 

%
%
%
\bibliographystyle{splncs04}
\bibliography{mybibliography}

\begin{thebibliography}{10}
\providecommand{\url}[1]{\texttt{#1}}
\providecommand{\urlprefix}{URL }
\providecommand{\doi}[1]{https://doi.org/#1}

\bibitem{Alberto2015}
Alberto, T.C., Lochter, J.V., Almeida, T.A.: {TubeSpam}: Comment spam filtering
  on {YouTube}. In: 2015 {IEEE} 14th International Conference on Machine
  Learning and Applications ({ICMLA}). {IEEE} (Dec 2015).
  \doi{10.1109/icmla.2015.37}, \url{https://doi.org/10.1109/icmla.2015.37}

\bibitem{Bach2017}
Bach, S.H., He, B., Ratner, A., R\'{e}, C.: Learning the structure of
  generative models without labeled data. In: Proceedings of the 34th
  International Conference on Machine Learning - Volume 70. p. 273–282.
  ICML'17, JMLR.org (2017)

\bibitem{barber2012}
Barber, D.: {Bayesian Reasoning and Machine Learning}. {Cambridge University
  Press} (2012)

\bibitem{Bartholomew1984}
Bartholomew, D.J.: The foundations of factor analysis. Biometrika
  \textbf{71}(2),  221--232 (1984). \doi{10.1093/biomet/71.2.221},
  \url{https://doi.org/10.1093/biomet/71.2.221}

\bibitem{bazavan2022}
Bazavan, E.G., Zanfir, A., Zanfir, M., Freeman, W.T., Sukthankar, R.,
  Sminchisescu, C.: Hspace: Synthetic parametric humans animated in complex
  environments (2022)

\bibitem{bishop2006}
Bishop, C.M.: Pattern Recognition and Machine Learning. Springer (2006)

\bibitem{Dai2016}
Dai, Z., Damianou, A., Gonzalez, J., Lawrence, N.D.: Variationally auto-encoded
  deep {G}aussian processes. In: Larochelle, H., Kingsbury, B., Bengio, S.
  (eds.) Proceedings of the International Conference on Learning
  Representations. vol.~3. Caribe Hotel, San Juan, PR (2016),
  \url{http://inverseprobability.com/publications/dai-variationally16.html}

\bibitem{Dunnmon2020}
Dunnmon, J.A., Ratner, A.J., Saab, K., Khandwala, N., Markert, M., Sagreiya,
  H., Goldman, R., Lee-Messer, C., Lungren, M.P., Rubin, D.L., R{\'{e}}, C.:
  Cross-modal data programming enables rapid medical machine learning. Patterns
   \textbf{1}(2),  100019 (May 2020). \doi{10.1016/j.patter.2020.100019},
  \url{https://doi.org/10.1016/j.patter.2020.100019}

\bibitem{Fabrigar1999}
Fabrigar, L.R., Wegener, D.T., MacCallum, R.C., Strahan, E.J.: Evaluating the
  use of exploratory factor analysis in psychological research. Psychological
  Methods  \textbf{4}(3),  272--299 (Sep 1999).
  \doi{10.1037/1082-989x.4.3.272},
  \url{https://doi.org/10.1037/1082-989x.4.3.272}

\bibitem{Fries2021}
Fries, J.A., Steinberg, E., Khattar, S., Fleming, S.L., Posada, J., Callahan,
  A., Shah, N.H.: Ontology-driven weak supervision for clinical entity
  classification in electronic health records. Nature Communications
  \textbf{12}(1) (Apr 2021). \doi{10.1038/s41467-021-22328-4},
  \url{https://doi.org/10.1038/s41467-021-22328-4}

\bibitem{goswami2022}
Goswami, M., Boecking, B., Dubrawski, A.: Weak supervision for affordable
  modeling of electrocardiogram data (2022)

\bibitem{jain2021}
Jain, N.: Customer sentiment analysis using weak supervision for customer-agent
  chat (2021)

\bibitem{Lawrence2006}
Lawrence, N.D., Qui{\~{n}}onero-Candela, J.: Local distance preservation in the
  {GP}-{LVM} through back constraints. In: Proceedings of the 23rd
  international conference on Machine learning - {ICML} {\textquotesingle}06.
  {ACM} Press (2006). \doi{10.1145/1143844.1143909},
  \url{https://doi.org/10.1145/1143844.1143909}

\bibitem{Liu2021}
Liu, Z., Zhu, X., Yang, L., Yan, X., Tang, M., Lei, Z., Zhu, G., Feng, X.,
  Wang, Y., Wang, J.: Multi-Initialization Optimization Network for Accurate 3D
  Human Pose and Shape Estimation, p. 1976–1984. Association for Computing
  Machinery, New York, NY, USA (2021),
  \url{https://doi.org/10.1145/3474085.3475355}

\bibitem{manco2021}
Manco, I., Benetos, E., Quinton, E., Fazekas, G.: Learning music audio
  representations via weak language supervision (2021)

\bibitem{Mathew2021}
Mathew, J., Negi, M., Vijjali, R., Sathyanarayana, J.: {DeFraudNet}: An
  end-to-end weak supervision framework to detect fraud in online food
  delivery. In: Machine Learning and Knowledge Discovery in Databases. Applied
  Data Science Track, pp. 85--99. Springer International Publishing (2021).
  \doi{10.1007/978-3-030-86514-6_6},
  \url{https://doi.org/10.1007/978-3-030-86514-6_6}

\bibitem{Murphy2012}
Murphy, K.P.: Machine Learning: A Probabilistic Perspective. MIT Press (2012)

\bibitem{rao2021}
Rao, V.R., Khalil, M.I., Li, H., Dai, P., Lu, J.: Decompose the sounds and
  pixels, recompose the events (2021)

\bibitem{Ratner2018b}
Ratner, A., Hancock, B., Dunnmon, J., Goldman, R., R\'{e}, C.: Snorkel metal:
  Weak supervision for multi-task learning. In: Proceedings of the Second
  Workshop on Data Management for End-To-End Machine Learning. DEEM'18,
  Association for Computing Machinery, New York, NY, USA (2018).
  \doi{10.1145/3209889.3209898}, \url{https://doi.org/10.1145/3209889.3209898}

\bibitem{Ratner2018a}
Ratner, A., Bach, S.H., Ehrenberg, H., Fries, J., Wu, S., R{\'{e}}, C.:
  Snorkel: rapid training data creation with weak supervision. The {VLDB}
  Journal  \textbf{29}(2-3),  709--730 (Jul 2019).
  \doi{10.1007/s00778-019-00552-1},
  \url{https://doi.org/10.1007/s00778-019-00552-1}

\bibitem{Ratner2019}
Ratner, A., Hancock, B., Dunnmon, J., Sala, F., Pandey, S., R{\'{e}}, C.:
  Training complex models with multi-task weak supervision. Proceedings of the
  {AAAI} Conference on Artificial Intelligence  \textbf{33},  4763--4771 (Jul
  2019). \doi{10.1609/aaai.v33i01.33014763},
  \url{https://doi.org/10.1609/aaai.v33i01.33014763}

\bibitem{Ratner2016}
Ratner, A.J., De~Sa, C.M., Wu, S., Selsam, D., R\'{e}, C.: Data programming:
  Creating large training sets, quickly. In: Lee, D., Sugiyama, M., Luxburg,
  U., Guyon, I., Garnett, R. (eds.) Advances in Neural Information Processing
  Systems. vol.~29. Curran Associates, Inc. (2016),
  \url{https://proceedings.neurips.cc/paper/2016/file/6709e8d64a5f47269ed5cea9f625f7ab-Paper.pdf}

\bibitem{reddy2021}
Reddy, R.G., Rui, X., Li, M., Lin, X., Wen, H., Cho, J., Huang, L., Bansal, M.,
  Sil, A., Chang, S.F., Schwing, A., Ji, H.: Mumuqa: Multimedia multi-hop news
  question answering via cross-media knowledge extraction and grounding (2021)

\bibitem{Saab2019}
Saab, K., Dunnmon, J., Goldman, R., Ratner, A., Sagreiya, H., R{\'{e}}, C.,
  Rubin, D.: Doubly weak supervision of deep learning models for head {CT}. In:
  Lecture Notes in Computer Science, pp. 811--819. Springer International
  Publishing (2019). \doi{10.1007/978-3-030-32248-9_90},
  \url{https://doi.org/10.1007/978-3-030-32248-9_90}

\bibitem{Saab2020}
Saab, K., Dunnmon, J., R{\'{e}}, C., Rubin, D., Lee-Messer, C.: Weak
  supervision as an efficient approach for automated seizure detection in
  electroencephalography. npj Digital Medicine  \textbf{3}(1) (Apr 2020).
  \doi{10.1038/s41746-020-0264-0},
  \url{https://doi.org/10.1038/s41746-020-0264-0}

\bibitem{Tipping1999}
Tipping, M.E., Bishop, C.M.: Probabilistic principal component analysis.
  JOURNAL OF THE ROYAL STATISTICAL SOCIETY, SERIES B  \textbf{61}(3),  611--622
  (1999)

\bibitem{tseng2021}
Tseng, A., Sun, J.J., Yue, Y.: Automatic synthesis of diverse weak supervision
  sources for behavior analysis (2021)

\bibitem{Varma2019}
Varma, P., Sala, F., He, A., Ratner, A., Re, C.: Learning dependency structures
  for weak supervision models. In: Chaudhuri, K., Salakhutdinov, R. (eds.)
  Proceedings of the 36th International Conference on Machine Learning.
  Proceedings of Machine Learning Research, vol.~97, pp. 6418--6427. PMLR
  (09--15 Jun 2019), \url{https://proceedings.mlr.press/v97/varma19a.html}

\bibitem{Wan2018}
Wan, M., McAuley, J.: Item recommendation on monotonic behavior chains. In:
  Proceedings of the 12th {ACM} Conference on Recommender Systems. {ACM} (Sep
  2018). \doi{10.1145/3240323.3240369},
  \url{https://doi.org/10.1145/3240323.3240369}

\bibitem{Wan2019}
Wan, M., Misra, R., Nakashole, N., McAuley, J.: Fine-grained spoiler detection
  from large-scale review corpora. In: Proceedings of the 57th Annual Meeting
  of the Association for Computational Linguistics. Association for
  Computational Linguistics (2019). \doi{10.18653/v1/p19-1248},
  \url{https://doi.org/10.18653/v1/p19-1248}

\bibitem{Weng2019}
Weng, Z., Varma, P., Masalov, A., Ota, J., Re, C.: Utilizing weak supervision
  to infer complex objects and situations in autonomous driving data. In: 2019
  {IEEE} Intelligent Vehicles Symposium ({IV}). {IEEE} (Jun 2019).
  \doi{10.1109/ivs.2019.8814147},
  \url{https://doi.org/10.1109/ivs.2019.8814147}

\bibitem{Weston2004}
Weston, J., Sch\"{o}lkopf, B., Bakir, G.: Learning to find pre-images. In:
  Thrun, S., Saul, L., Sch\"{o}lkopf, B. (eds.) Advances in Neural Information
  Processing Systems. vol.~16. MIT Press (2004),
  \url{https://proceedings.neurips.cc/paper/2003/file/ac1ad983e08ad3304a97e147f522747e-Paper.pdf}

\bibitem{wolfson2021}
Wolfson, T., Berant, J., Deutch, D.: Weakly supervised mapping of natural
  language to sql through question decomposition (2021)

\bibitem{zhang2021}
Zhang, J., Yu, Y., Li, Y., Wang, Y., Yang, Y., Yang, M., Ratner, A.: {WRENCH}:
  A comprehensive benchmark for weak supervision. In: Thirty-fifth Conference
  on Neural Information Processing Systems Datasets and Benchmarks Track
  (2021), \url{https://openreview.net/forum?id=Q9SKS5k8io}

\bibitem{zheng2021}
Zheng, J., Shi, X., Gorban, A., Mao, J., Song, Y., Qi, C.R., Liu, T., Chari,
  V., Cornman, A., Zhou, Y., Li, C., Anguelov, D.: Multi-modal 3d human pose
  estimation with 2d weak supervision in autonomous driving (2021)

\end{thebibliography}

\section*{Disclaimer}

This paper was prepared for informational purposes by the Applied Innovation of AI (AI2) and Global Technology Applied Research center of JPMorgan Chase \& Co. This paper is not a product of the Research Department of JPMorgan Chase \& Co. or its affiliates. Neither JPMorgan Chase \& Co. nor any of its affiliates makes any explicit or implied representation or warranty and none of them accept any liability in connection with this paper, including, without limitation, with respect to the completeness, accuracy, or reliability of the information contained herein and the potential legal, compliance, tax, or accounting effects thereof. This document is not intended as investment research or investment advice, or as a recommendation, offer, or solicitation for the purchase or sale of any security, financial instrument, financial product or service, or to be used in any way for evaluating the merits of participating in any transaction.

\end{document}